\documentclass{svproc}
\usepackage{url}

\usepackage{graphicx}
\usepackage[bookmarks=true]{hyperref}
\usepackage{subcaption}
\usepackage{cleveref} 
\makeatletter
\def\thanks#1{\protected@xdef\@thanks{\@thanks
        \protect\footnotetext{#1}}}
\makeatother
\graphicspath{{figures/}}

\begin{document}
\mainmatter              
\title{Ordered Genetic Algorithm for Entrance Dependent Vehicle Routing Problem in Farms}
\titlerunning{OGA for EVRP in farms}  
%
\author{Haotian Xu *\inst{1} \and Xiaohui Fan *\inst{1} \and Jialin Zhu\inst{2} \and Qing Zhuo\inst{1} \and Tao Zhang\inst{1}$^{,\dag}$
\thanks{* equal contribution.}
\thanks{$\dag$ Corresponding author.}
\thanks{This research was funded by National Science and Technology Major Project under Grant 2021ZD0110900.}
}
\authorrunning{Haotian Xu et al.} 
%
%
\institute{Department of Automation, Tsinghua University, Beijing, China,\\
\email{taozhang@tsinghua.edu.cn} \\
\and China Academy of Launch Vehicle Technology, Beijing, China
}
\maketitle              

\begin{abstract}
Vehicle Routing Problems (VRP) are widely studied issues that play important roles in many production scenarios. We have noticed that in some practical scenarios of VRP, the size of cities and their entrances can significantly influence the optimization process. To address this, we have constructed the Entrance Dependent VRP (EDVRP) to describe such problems. We provide a mathematical formulation for the EDVRP in farms and propose an Ordered Genetic Algorithm (OGA) to solve it. The effectiveness of OGA is demonstrated through our experiments, which involve a multitude of randomly generated cases. The results indicate that OGA offers certain advantages compared to a random strategy baseline and a genetic algorithm without ordering. Furthermore, the novel operators introduced in this paper have been validated through ablation experiments, proving their effectiveness in enhancing the performance of the algorithm.

\end{abstract}
\section{Introduction}
Vehicle Routing Problem (VRP) is a combinatorial optimization issue, widely applied across industries, logistics, postal services, and even integrated into chip design. VRP addresses the strategic deployment of vehicles, constituting a critical component in guidance and navigation systems for route optimization.

In the VRP, multiple vehicles need to traverse several cities using the shortest paths without omission or repetition. There are various variations of VRP that arise based on different application contexts. CVRP (Capacitated VRP) \cite{RN95,RN94} is a widely studied variation of VRP that is based on the delivery of goods by trucks. It introduces constraints on truck capacity and city demand, transforming the problem into how multiple trucks with limited cargo capacity can travel the shortest paths to fulfill the demands of all cities. In some cases, there are constraints on tasks or demands regarding their earliest or latest completion times, leading to VRP with Time Window (VRPTW) \cite{RN96,RN97}. Some researchers point out that the running speed of vehicles in real production may not remain constant, so they introduce Time Dependent VRP (TDVRP) \cite{RN92,RN93,RN91}. In actual problems, multiple depots lead to Multi-Depot VRP (MDVRP) \cite{RN98,RN100}, and the various types of trucks lead to Heterogeneous Fleet VRP (HFVRP) \cite{RN102,RN14}. 

However, Previous work often simplifies cities as points, ignoring their size's impact. In reality, VRP nodes, including cities, vary in size and have multiple entrances. Clearly, vehicle distances change with different city entrances. We call this VRP variant as Entrance Dependent VRP (EDVRP). An algorithm must select entrances for each city in EDVRP.

\begin{figure}[t]
    \centering
    \includegraphics[width=0.48\linewidth]{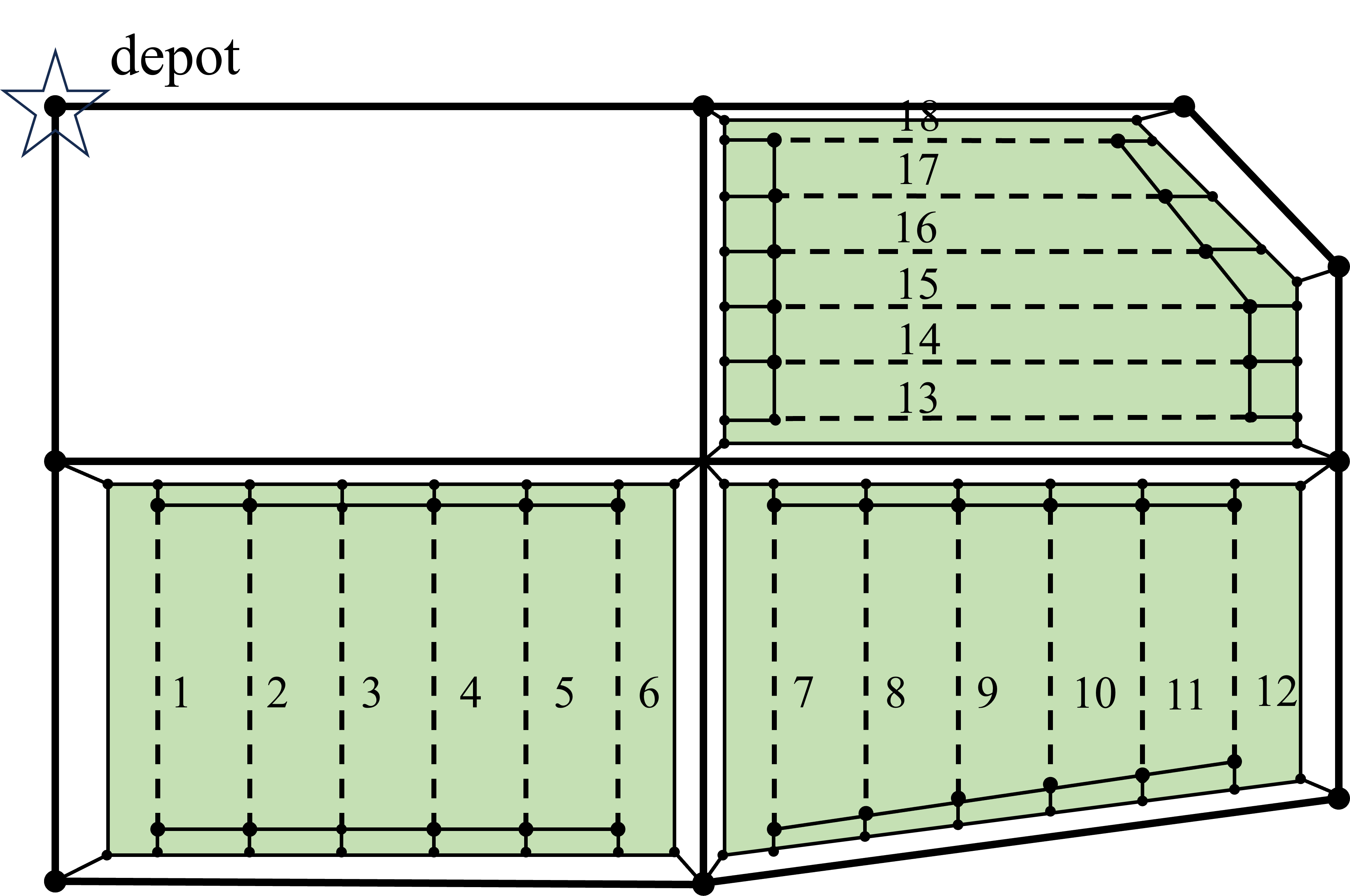}
    \caption{A typical farm. The green polygons represent three fields, while the pentagrams denote the garage(depot). 
    Thick solid lines represent roads within the farm, thin solid lines depict passable roads within fields, and dashed lines indicate working lines awaiting operation. 
    The numbers on the edges of the working lines are their identifiers.}
    \label{fig:Oc_topo}
\end{figure}

EDVRP can be applied in various scenarios. In this paper, we focus on arranging agricultural machinery in farms, illustrated in Fig. \ref{fig:Oc_topo}. Dashed lines denote operational needs like planting, spraying, plowing. 'Working lines' (dashed) represent EDVRP's nodes or 'cities'. Each node has two entrances: a machine enters one and leaves through the other.   

Heuristic search algorithms are commonly used to solve VRP, including the Greedy Search Algorithm \cite{RN12}, Backtracking Search Algorithm (BSA) \cite{RN95,RN61}, Ant Colony Optimization (ACO) \cite{RN97,RN62}, Simulated Annealing (SA) \cite{RN115}, Genetic Algorithm (GA) \cite{RN96,RN92,RN16}, Particle Swarm Optimization (PSO) \cite{RN112}, Tabu Search (TS) \cite{RN94,RN93,RN97}, and others. 
However, to the best of our knowledge, no existing algorithm focuses on EDVRP. 
In this paper, expanding upon the multi-variant genetic algorithm proposed by Wang Meng et al. \cite{RN16}, we introduce entrance inversion, intra-group sorting, and greedy path operators to extend the genetic algorithm into an Ordered Genetic Algorithm (OGA) to solve the EDVRP in farms. Our contribution can be summarized as follows:
\begin{itemize}
    \item We derived the Entrance Dependent VRP (EDVRP), to descirbe the senario where cites' size and entrance influence the distance or time or other metrics.
    \item We formulated the EDVRP in farms, and proposed Ordered Genetic Algorithm to solve it, providing an experiments-validated baseline for EDVRP. 
\end{itemize} 

\section{Methodology}
\subsection{Problem Fomulation}
In this paper, the depot and working lines are considered as nodes, 
forming a task graph $G$. 
Let the number of working lines be denoted by $L$. The set of nodes on the graph is denoted by $N=\{\mathbf{n}_i|i=0,1,...,L\}$, 
where $\mathbf{n}_0$ represents the depot, and $\mathbf{n}_1$ to $\mathbf{n}_L$ represent working lines. 
We use $l_i$ to represent the length of the working line $i$. 
For the depot, $l_0=0$. 

There is an edge connecting any two different nodes in the graph.
Since each working line has two entrances, there are four distinct distances between any two nodes 
$\mathbf{n}_i$ and $\mathbf{n}_j$. 
We denote by $m_i \in \{0,1\}$ the entrance selection for the $i$-th working line. 
In this paper, we define $d^*_{ij m_i m_j}$ to represent the shortest traversable distance between the entrance $m_i$ of working line $\mathbf{n}_i$ 
and the entrance $m_j$ of working line $\mathbf{n}_j$. 
For consistency, we assume that the depot also has two entrances which have the same coordinates. 
The features of the edges in the task graph are given by $\mathbf{d}_{ij}=\left[ d^*_{ij00}, d^*_{ij01}, d^*_{ij10}, d^*_{ij11} \right]$, 
and the set of edges is denoted as $E=\{\mathbf{d}_{ij}|i=0,1,...,L;j=0,1,...,L;i \neq j\}$.

Considering the variations of machines, the parameters of the $k$-th machine are defined by its working speed $v_k^w$, 
idle speed $v_k^v$, fuel consumption during operation $c_k^w$, and fuel consumption during idling $c_k^v$, 
denoted as $\mathbf{m}_k=\left[v_k^w,v_k^v,c_k^w,c_k^v\right]$. 
All distances are measured in meters (m), time is measured in seconds (s), and fuel consumption is measured in liters (L).

The optimization variable is a permutation $P$ of all working lines and their corresponding chosen entrances. 
Each working line appears exactly once in the permutation, indicating that each working line must be serviced exactly once. 
Let there be a total of $M$ machines, then the permutation is divided into $M$ segments, each representing the sequence of tasks for one machine. 
We denote $P_{k,i}$ as the $i$-th working line assigned to the $k$-th machine's route, and $n_k$ as the sequence length of $P_k$. 
The chosen entrance of $P_{k,i}$ is denoted as $e_{k,i}$, and the other entrance (which is actualy the exit) is denoted as $\bar{e}_{k,i}$. 
Therefore, the idle distance for the $k$-th machine is
\begin{equation}
    s_k=d^*_{0 P_{k,0} 0 e_{k,1}}+d^*_{P_{k,n_k} 0 \bar{e}_{k,n_k} 0}+\sum_{i = 0}^{n_k-1} d^*_{P_{k,i} P_{k,i+1} \bar{e}_{k,i} e_{k,i+1}}.
    \label{eq:idle_len}
\end{equation}
The first term in equation (\ref{eq:idle_len}) represents the distance from the depot to the first working line of the $k$-th machine, 
the second term represents the distance from the last working line of the $k$-th machine back to the depot, 
and the third term is the sum of distances for the $k$-th machine between consecutive working lines.
The operation distance for the $k$-th machine is
\begin{equation}
    s_k^w=\sum_{i=0}^{n_k} l_{P_{k,i}}.
\end{equation}
Its total time cost and total fuel consumption thus can be calculated as
\begin{equation}
    t_k=\frac{s_k}{v_k^v}+\frac{s_k^w}{v_k^w}, \quad c_k^o=\frac{s_k}{v_k^v}c_k^v+\frac{s_k^w}{v_k^w}c_k^w.
\end{equation}
 
Since the total operation distance for all machines is a constant, we use total idle distance $s_P$, total time $t_P$ and total fuel consumption $c_P$ as objectives:
\begin{equation}
    s_P=\sum_{k=1}^{M}s_k, \quad t_P=\max_k t_k, \quad c_P=\sum_{k=1}^{M}c_k^o.
    \label{eq:theorm_sp}
\end{equation}

    

   

\subsection{Ordered Genetic Algorithm}
In genetic algorithms, a solution to the problem is called as a chromosome, 
and chromosomes exist within populations. 
Following the research of Wang Meng et al. \cite{RN16}, 
we divide the chromosome into two parts: 
the first part consists of a permutation $T$ of all working lines, 
and the second part comprises a set of segment points that divide $T$ into $M$ segments. 
Each element of $T$ is a tuple consisting of the index $P_i$ of a working line and the corresponding chosen entrance $e_i$, 
denoted by $T_i=[P_i,e_i]$.

In genetic algorithms, individuals with higher fitness values are more likely to advance to the next iteration of the solution process. The fitness we calculate is the reciprocal of the objective values. A higher fitness value of a chromosome indicates that the arrangement it represents is of higher quality. Specifically, for the three objectives, the fitness values are calculated as the reciprocals of the respective objective values, given by $1/s_P$, $1/t_P$, and $1/c_P$, respectively.

Genetic algorithms consist of three basic operations: selection, crossover, and mutation. Genetic algorithms iteratively obtain optimal individuals by repeatedly applying these operations to a population with a defined size, terminating when the iterations reach the maximun.

The selection operator samples fit individuals as parents using the roulette wheel method where an individual's fitness is it's probability of being chosen. The crossover operator creates new offspring by exchanging parts of two parents, as detailed in \cite{RN16}. It selects unique task sequences from each parent to form the offspring, removing any duplicate working lines. Unfilled positions are randomly filled with remaining lines. The fittest parent replaces the weakest offspring if superior, ensuring the best traits are retained. The output is the fittest individual when the algorithem terminates.

The mutation operator of genetic algorithms randomly alters individuals with a certain probability. The mutation operators in OGA include 3 commonly used operators from \cite{RN16}: the local inversion, inter-group exchange and the inter-group transfer operator and 3 noval operators proposed by us: the entrance inversion, the intra-group sorting, and the greedy path operator. 

The local inversion operator reverses a segment of the sequence $T$. 
The inter-group exchange operator swaps two working lines within the task sequences of two machines. The inter-group transfer operator moves a working line from one task sequence to another. These operators do not take entrances into account.

Considering the impact of entrances, it is necessary to explore the entrances for each working line. The entrance inversion operator randomly selects one machine and reverses the entrances of all working lines in its task sequence. As the crossover and mutation operators act between different machines' sequences, this reversal operation will have an impact on the overall arrangement.

\begin{figure}
    \centering
    \includegraphics[width=0.25\linewidth]{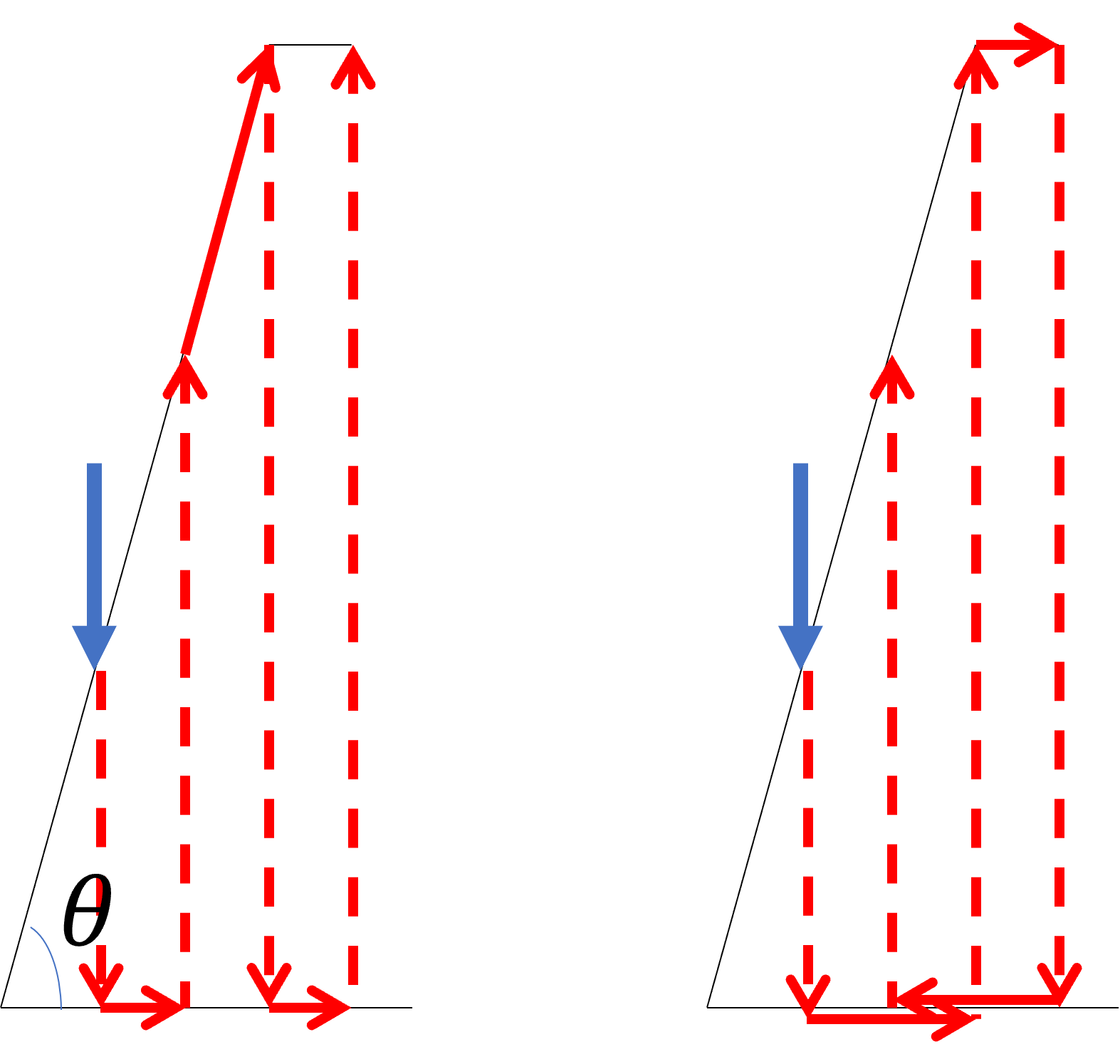}
    \caption{In the scenario, the working lines represented by the red dashed lines have been assigned to one machine, which enters the field from above the leftmost working line. 
    It can be observed that when $\cos \theta < 1/3$, the distance traveled for the left sequence is longer than that for the non-sequential operation on the right.}
    \label{fig:ant_order}
\end{figure}

Working lines within a single field are actually in a certain order. In a single field, working from one side to the other is likely to be more efficient. Therefore, we have designed an intra-group sorting operator to sort the working lines within the same field for the same machine and correspondingly alternate the entrances of these working lines. However, as illustrated by the counterexample in Fig. \ref{fig:ant_order}, simply working in sequence does not guarantee an optimal path.

\begin{figure}
    \centering
    \includegraphics[width=0.25\linewidth]{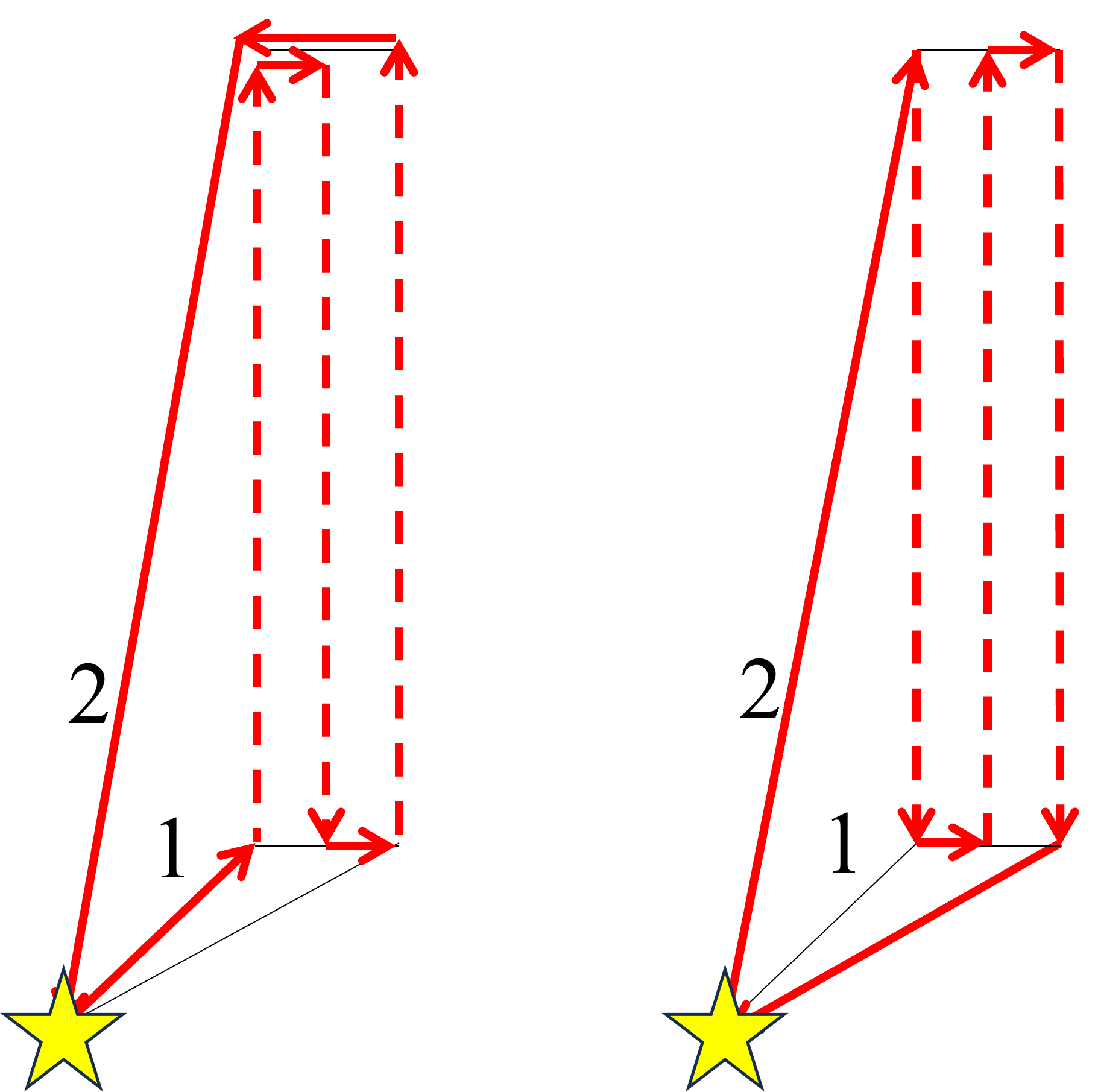}
    \caption{The machine needs to complete tasks on three working lines from left to right. 
    The left solution selects path 1 from the depot to the leftmost working line based on the greedy operator, 
    while the right solution chooses the longer path 2. 
    It is visually evident that the greedy path on the left is longer than the one on the right.}
    \label{fig:ant_greedy}
\end{figure}

Solving EDVRP, a direct approach is to select the shortest path among the traversable paths between each pair of nodes when the sequence of nodes is determined. However, because of the lengths of the working lines and the constraint that each working line can only be serviced once, there is a restriction that, once the chosen entrance $e_i$ of working line $i$ is determined, its exit $\bar{e}_i$ is also determined, 
thus limiting the path choices between working line $i$ and the subsequent working line $j$ to either $d^*_{ij\bar{e}_i 0}$ or $d^*_{ij\bar{e}_i 1}$. 
Therefore, we design a greedy path operator, which, for each machine's task sequence, selects the shortest path among the two feasible paths between every two adjacent working lines, starting from the depot to the first working line. It is worth noting that the selected path may not necessarily be the shortest among the four paths between adjacent working lines, and the path from the last working line to the depot is determined when the entrance of the last working line is determined. Therefore, the greedy path may not be the optimal solution either. Fig. \ref{fig:ant_greedy} provides a counterexample.

The entrance inversion, intra-group sorting, and greedy path operator operate internally on the task sequences of individual machines. The term 'ordered' in the context of OGA refers to the utilization of both the intra-group sorting operator and the greedy path operator, which consider the order of the working lines and perform sequential greedy searches.

\begin{figure}
    \centering
    \includegraphics[width=0.8\linewidth]{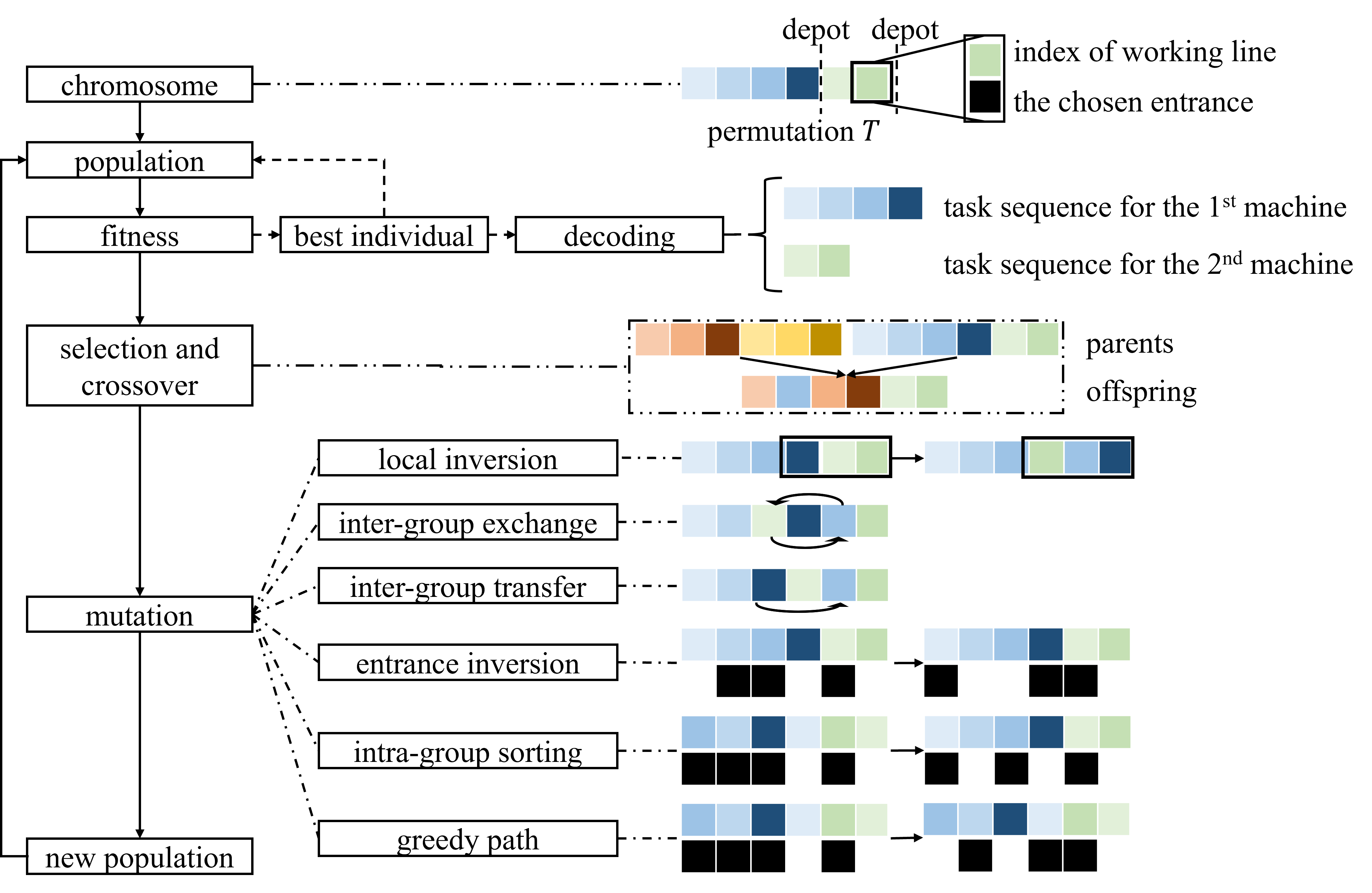}
    \caption{Flowchart and visualization of operators.}
    \label{fig:GA}
\end{figure}

The flowchart of OGA and operators in OGA are illustrated in Fig. \ref{fig:GA}
\section{Experiments}
\subsection{Compare with Baselines}
We construct experiments using randomly generated datasets to demonstrate the effectiveness and strengths of OGA. The datasets include 4500 cases. Each case has a farm with 1, 2, 3, 4, or 6 fields and 3, 5, or 7 machines. Each farm consists of approximately 40 nodes, comprising working lines and a depot.

In the experiments, OGA's population size is 200, the number of iterations is 200, the probabilities of crossover, local inversion, inter-group exchange, inter-group transfer, entrance inversion, intra-group sorting, and greedy path operators are 0.6, 0.5, 0.5, 0.6, 0.5, 0.8, and 0.8, respectively. The results are presented in Table \ref{tab:GA_test_s}. The baseline algorithm used for comparison is "rand", which generates arrangements randomly within a 1-second timeframe and selects the best arrangement as the output. The results demonstrate that OGA effectively optimizes the EDVRP for farms, leading to cost reduction and improved efficiency.
\begin{table}
    \centering
    \caption{Test results. The letters 's', 't' and 'c' in bracket means the algorithm use $s_p$, $t_P$ and $c_P$ as objective, respectively}
    \begin{tabular}{lccc}
        \hline
        \rule{0pt}{12pt}
                &   Average $s_P$(m)  &  Average $t_P$(s)  &  Average $c_P$(L)  \\[2pt]
        \hline\rule{0pt}{12pt}

        rand(s)   &      19243.17       &     6555.80        &     98.99          \\
        OGA(s)    & \textbf{7153.59}    &     5603.64        &     79.78          \\
        rand(t)   &        24658.77     &      3284.49       &    100.57 \\
        OGA(t)    &        11521.66     &  \textbf{2306.18}  &    80.24      \\
        rand(c)   &       22587.59      &       7603.39      &     79.39        \\
        OGA(c)    &       8285.69       &      5518.89       &     \textbf{59.03}        \\[2pt]
        \hline
    \end{tabular}
    \label{tab:GA_test_s}
\end{table}



\subsection{Case Study}
\begin{figure}
    \centering
    \includegraphics[width=0.35\linewidth]{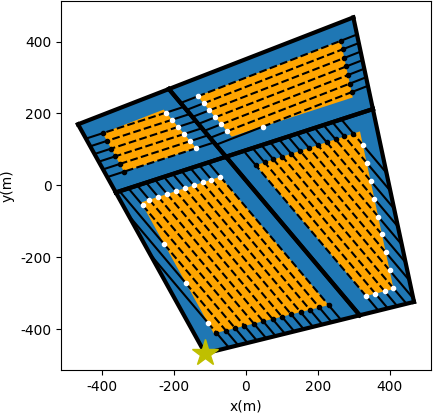}
    \caption{Farm used in case study. The star represents the depot. The dashed lines are the working lines. 
    The white and black points are the entrances 0 and 1, respectively.}
    \label{fig:render1}
\end{figure}

We analyze the detailed results through the following test case. 
The farm configuration for the test case is illustrated in Fig. \ref{fig:render1}. 

Four machines are utilized, whose parameters are detailed in Table \ref{tab:test_car_cfg}. M4 possesses the highest speed but also the highest fuel consumption per unit distance, while M2 exhibits the lowest fuel consumption per unit distance.
\begin{table}[h!]
    \centering
    \caption{Parameters of machines}
    \begin{tabular}{lcccc}
        \hline
        \rule{0pt}{12pt}
           &       $v^w$(m/s)     &        $v^v$(m/s)   &      $c^w$(L/s)    &    $c^v$(L/s) \\[2pt]
           \hline
           \rule{0pt}{12pt}
      M1   &         2.5          &          4.5        &       0.008        &      0.006          \\
      M2   &         3            &          5          &       0.007        &      0.005          \\
      M3   &         3            &          5.5        &       0.008        &      0.006          \\
      M4   &         3            &          6          &       0.01         &      0.008          \\[2pt]
      \hline
    \end{tabular}
    \label{tab:test_car_cfg}
\end{table}

\begin{figure}[h!]
    \centering 
    Optimization process

      \includegraphics[width=0.32\linewidth]{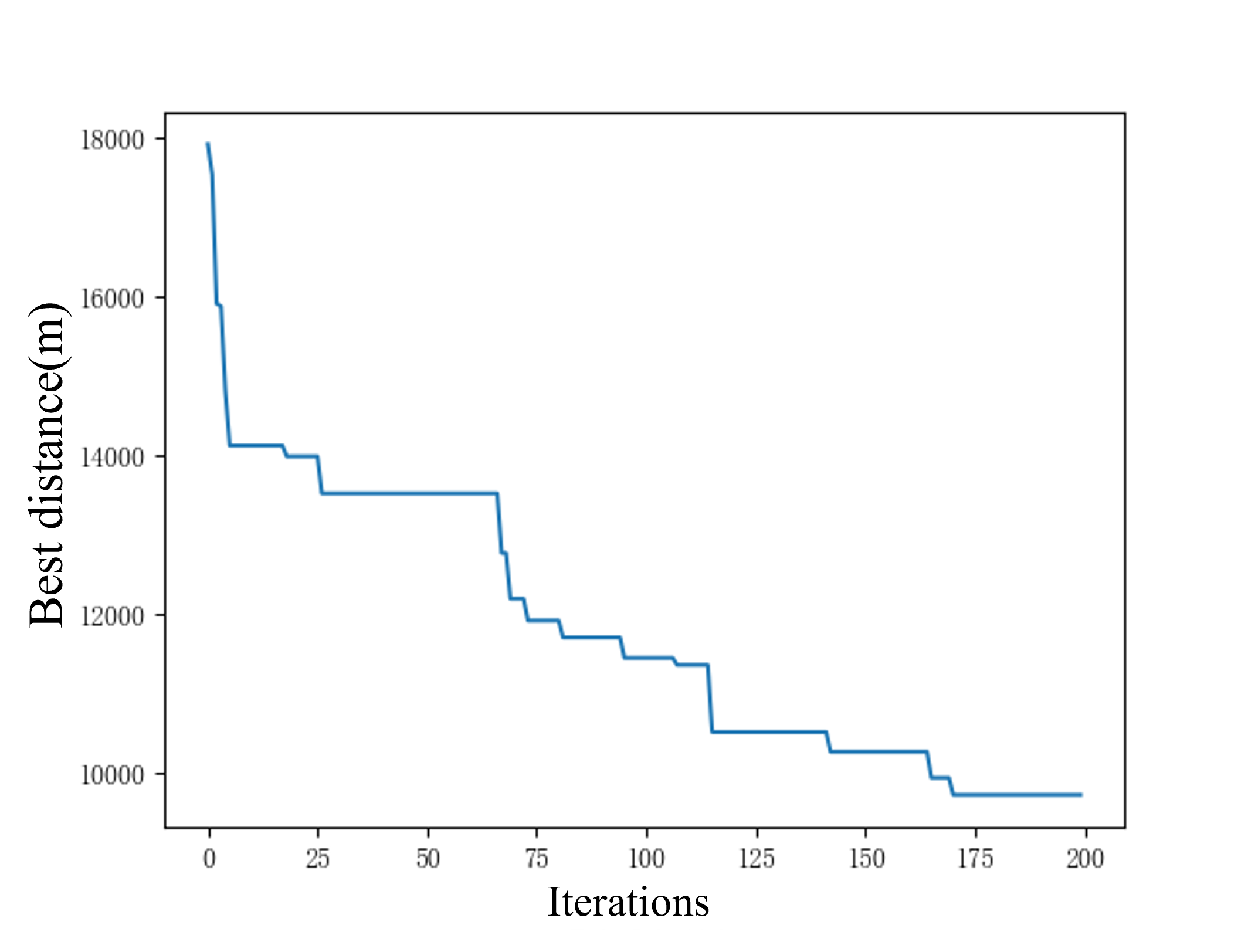}
      \includegraphics[width=0.32\linewidth]{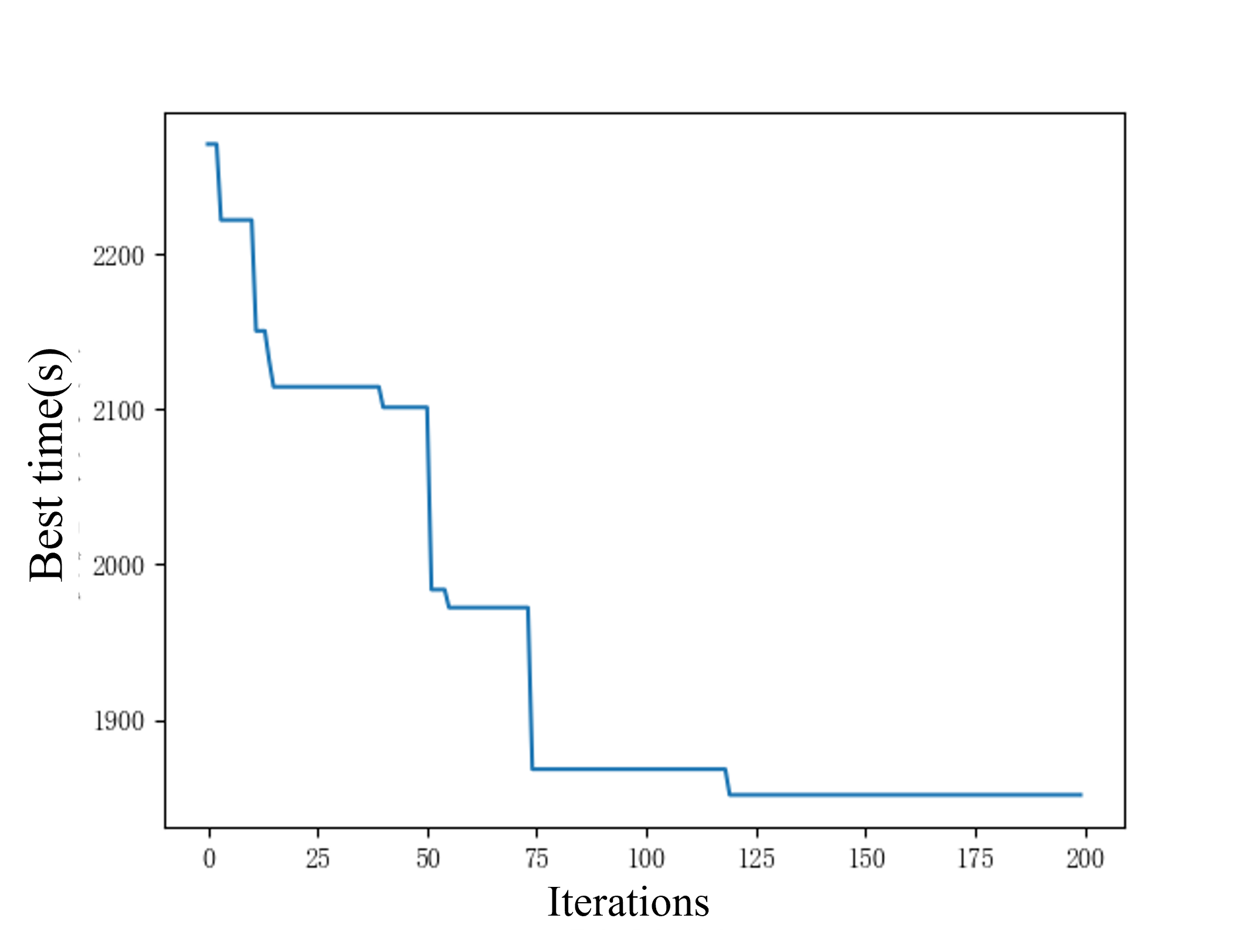}
      \includegraphics[width=0.32\linewidth]{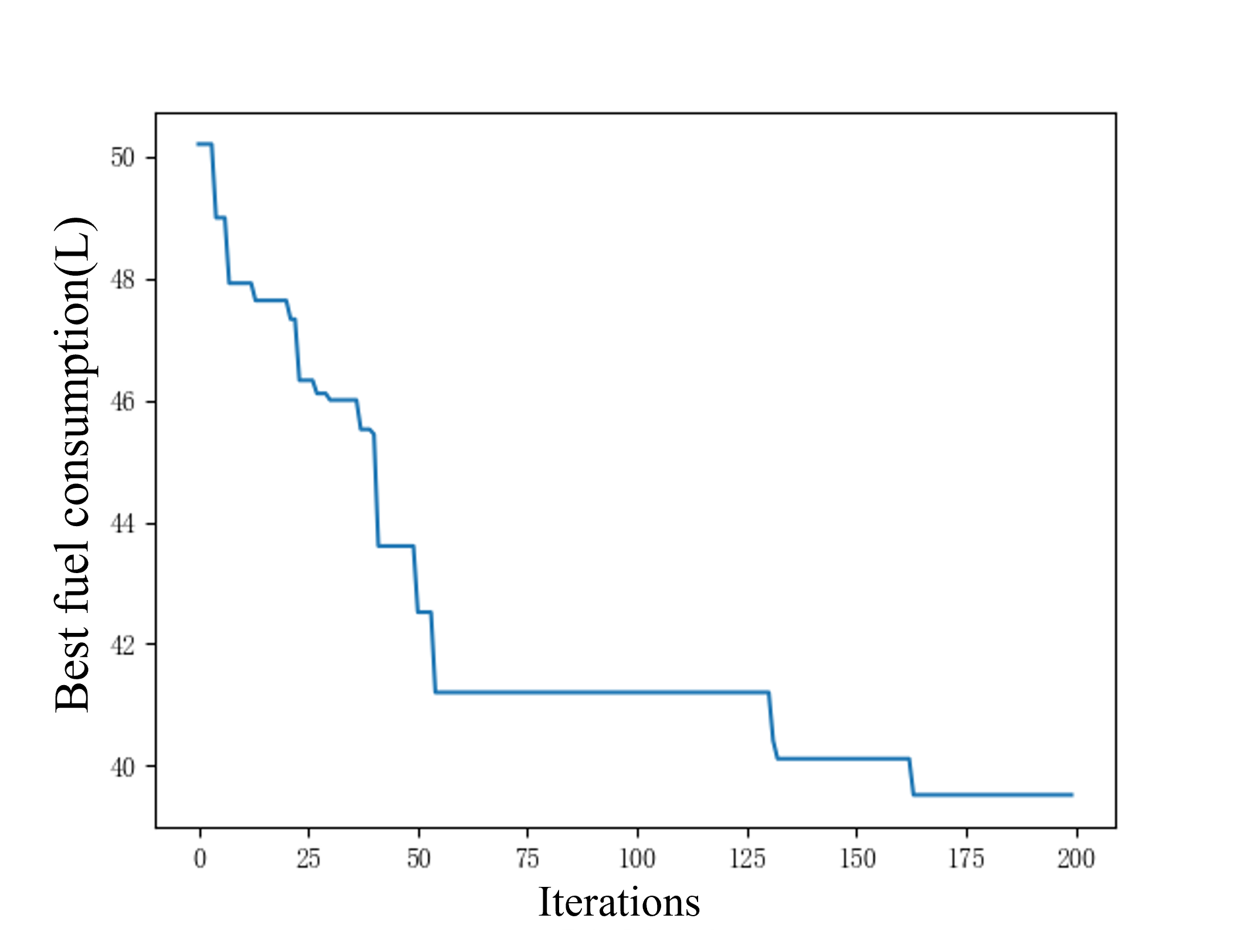}
    Optimization results
    
    \begin{subfigure}{0.32\linewidth}
        \includegraphics[width=\linewidth]{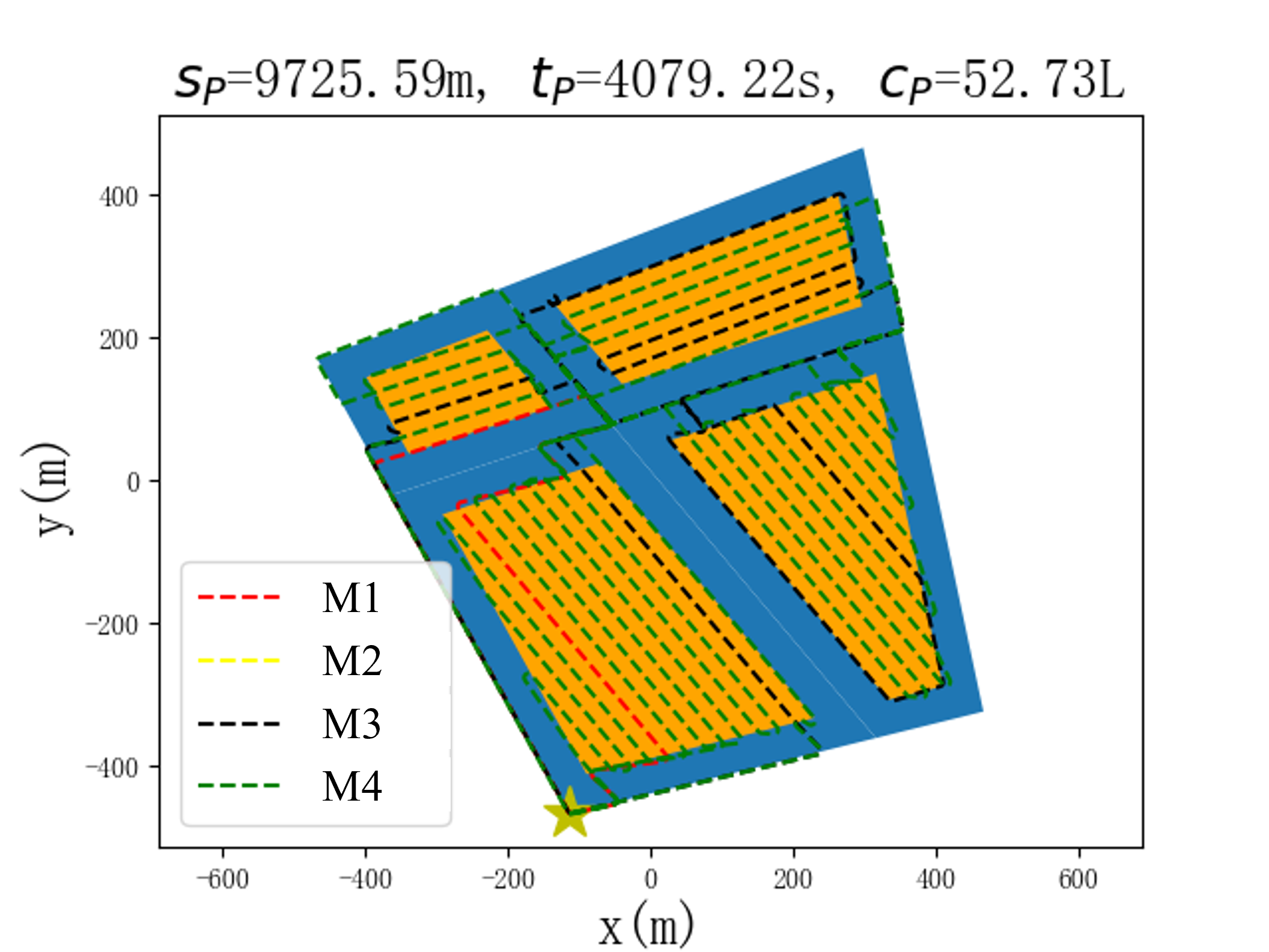}
        \caption{Optimizing $s_P$ \label{fig:GA_s}}
    \end{subfigure}
    \begin{subfigure}{0.32\linewidth}
        \includegraphics[width=\linewidth]{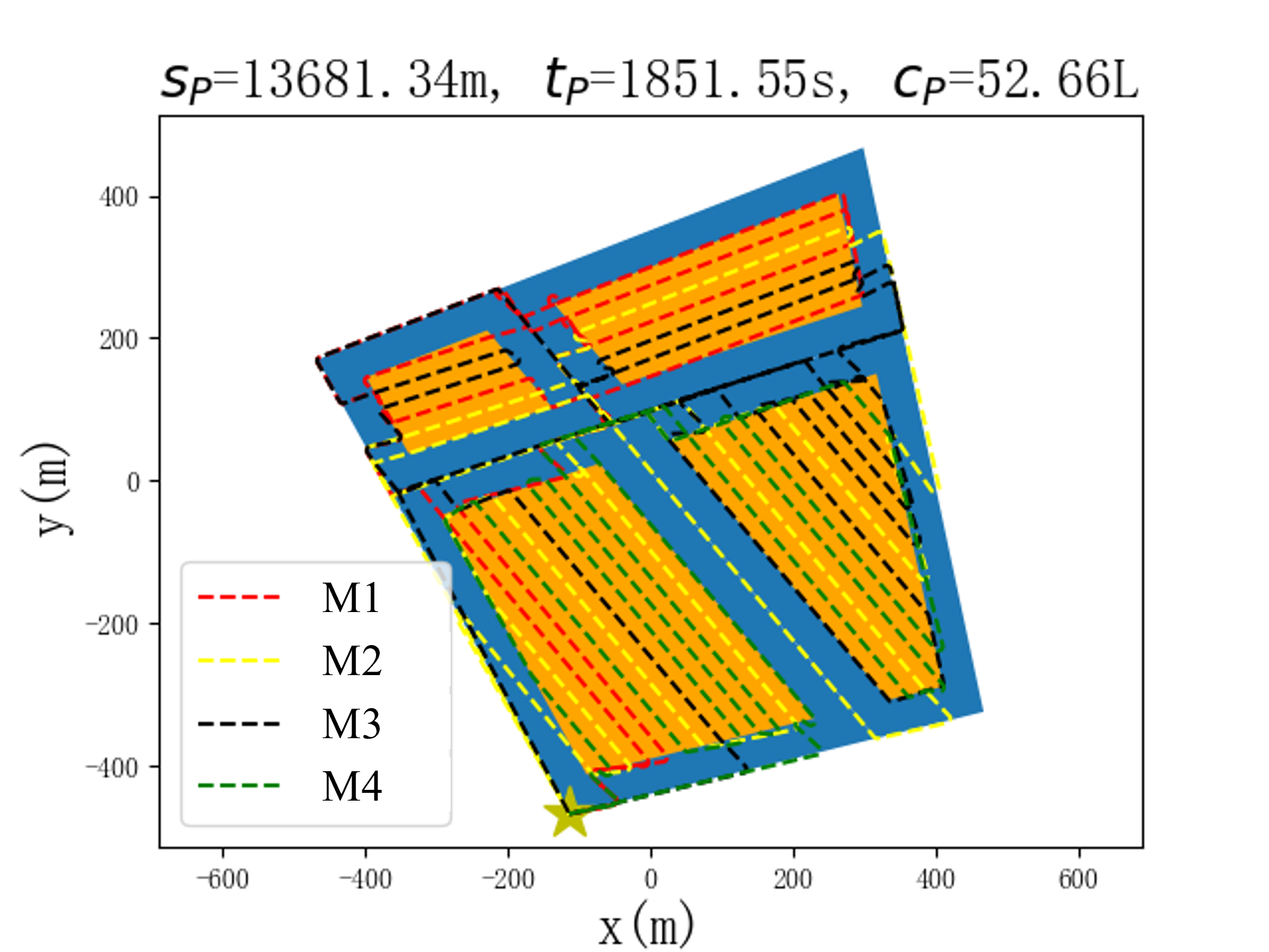}
        \caption{Optimizing $t_P$ }
    \end{subfigure}
    \begin{subfigure}{0.32\linewidth}
        \includegraphics[width=\linewidth]{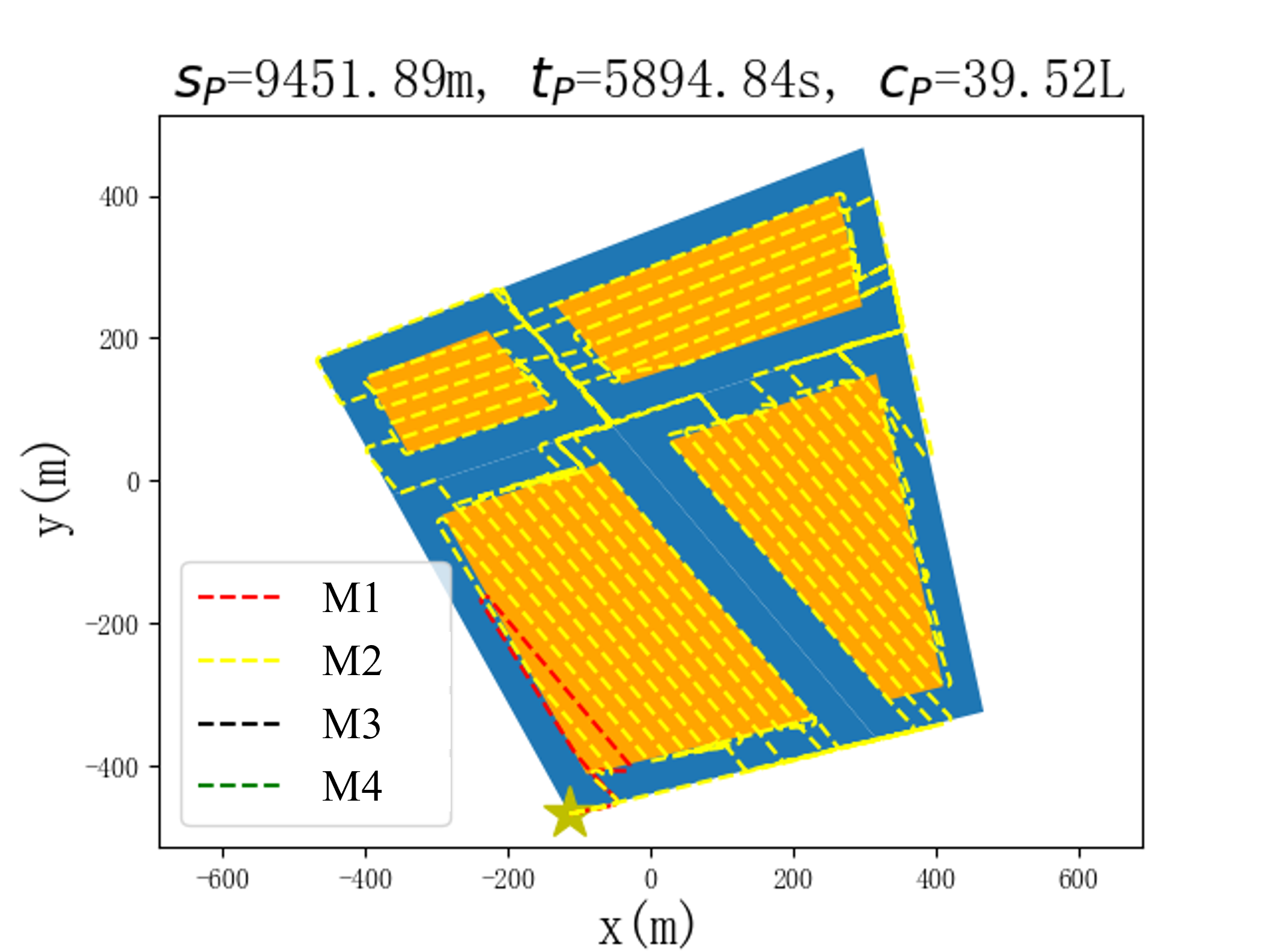}
        \caption{Optimizing $c_P$ \label{fig:GA_c}}
    \end{subfigure}
    \caption{The optimization process shows the variation of the objective values of the best individual in the population over the iterations. 
    The optimization results display the solutions, represented as machines' paths on the farm.}
    \label{fig:GA_case_study}
\end{figure}

The optimization process and results of OGA for three objectives are presented in Fig. \ref{fig:GA_case_study}. According to results, minimizing total distance leads to tasks being mainly assigned to one machine. Fig. \ref{fig:GA_s} shows machines traverse shorter distances along boundaries and roads. The distance traveled is machine parameter-independent, so tasks may be allocated to a machine with higher fuel consumption, as shown in Fig. \ref{fig:GA_s}. Minimizing total time prompts the algorithm to distribute tasks across all machines, with faster ones getting more, ensuring balanced workload and less idle time. When minimizing fuel consumption, the algorithm picks M2, the machine with lowest fuel consumption, for most tasks. This approach also results in a short total distance, as compared to Fig. \ref{fig:GA_s}.

\subsection{Ablation}
We conduct ablation tests to assess the impact of intra-group sorting and greedy path operators. We label the algorithm using greedy paths alone as OGA-greedy, the one with intra-group sorting only as OGA-sort, and the one lacking both as GA-no-order. GA-no-order is actually the MGGA from \cite{RN16} attached by entrance inversion operator.

We create a small dataset for ablation experiments. We sample sixty cases for each combination of field and machine numbers, yielding a total of 900 cases. The results of these ablation experiments are presented in Table \ref{tab:GA_ablation_s}.

\begin{table}
    \centering
    \caption{Ablation results}
    \begin{tabular}{lccc}
        \hline
        \rule{0pt}{12pt}
                         &  Average $s_P$(m) &  Average $t_P$(s)&  Average $c_P$(L)  \\[2pt]
        \hline\rule{0pt}{12pt}
      OGA(s)             & \textbf{7368.00}  &    5276.69       &    77.43         \\
      OGA-greedy(s)      &     9603.01       &    4605.85       &    82.49         \\
      OGA-sort(s)        &     8764.86       &    5550.62       &    80.59         \\
      GA-no-order(s)     &     15757.63      &    5550.78       &    92.31         \\
      OGA(t)             &     11690.76      & \textbf{2279.35} &    78.98      \\
      OGA-greedy(t)      &     13329.34      &    2375.57       &    81.61         \\
      OGA-sort(t)        &     14070.89      &    2410.10       &    82.58         \\
      GA-no-order(t)     &     22212.51      &    2862.91       &    95.07         \\
      OGA(c)             &     8644.58       &    5444.86       &    \textbf{58.09}        \\
      OGA-greedy(c)      &     10854.19      &    5730.25       &    61.22         \\
      OGA-sort(c)        &     10314.39      &    5652.06       &    60.52         \\
      GA-no-order(c)     &     18079.81      &    6814.55       &    70.54         \\[2pt]
      \hline
    \end{tabular}
    \label{tab:GA_ablation_s}
\end{table}


The ablation results show that the intra-group sorting and the greedy path
operator effectively enhancing the algorithm’s
performance compared to the unordered GA-no-order across various optimization objectives. The operators optimize the task sequence in two different aspects. The intra-group sorting operator optimizes the task sequence within a
field under the premise of determining the entrance direction of the field, while
the greedy path operator sequentially selects the shortest feasible path based
on the overall task sequence. The results of OGA-sort and OGA-greedy are
comparable. However, when combined, the algorithm’s performance is further
improved. OGA, which utilizes both the 2 operators, outperforms algorithms
that only adopt one of these operators across all three objectives.

\section{Conclusion}
In this paper, we notice that in some practical scenarios of VRP, the cites’ size and
entrance can significantly influence the optimization, and construct EDVRP to
describe this kind of problems. We give mathematical formulation of an EDVRP
scenario, which is machines allocation in farm. By proposing chromosome encoding considering entrance and entrance inversion, intra-group sorting and greedy
path operator, we improve genetic algorithm to OGA, to solve the EDVRP in
farm. Experiments proves the effectiveness and benefits of OGA, compared with
baselines. In the future, we hope OGA can be a strong baseline for EDVRP in more scenarios
besides farms, and the usage of neural networks is promising future work.

\bibliographystyle{unsrt}
\bibliography{ref_lib}

\begin{thebibliography}{10}

\bibitem{RN95}
Mahmuda Akhtar, MA~Hannan, RA~Begum, Hassan Basri, and Edgar Scavino.
\newblock Backtracking search algorithm in cvrp models for efficient solid waste collection and route optimization.
\newblock {\em Waste Management}, 61:117--128, 2017.

\bibitem{RN94}
Philippe Augerat, José-Manuel Belenguer, Enrique Benavent, Angel Corbéran, and Denis Naddef.
\newblock Separating capacity constraints in the cvrp using tabu search.
\newblock {\em European Journal of Operational Research}, 106(2-3):546--557, 1998.

\bibitem{RN96}
T~Gocken and M~Yaktubay.
\newblock Comparison of different clustering algorithms via genetic algorithm for vrptw.
\newblock {\em International Journal of Simulation Modeling}, 18(4), 2019.

\bibitem{RN97}
Vitoria Pureza, Reinaldo Morabito, and Marc Reimann.
\newblock Vehicle routing with multiple deliverymen: Modeling and heuristic approaches for the vrptw.
\newblock {\em European Journal of Operational Research}, 218(3):636--647, 2012.

\bibitem{RN92}
Mehmet~Selçuk Korkmaz.
\newblock {\em Genetic Algorithm Based Solution Approach for TDVRP}.
\newblock Thesis, Marmara Universitesi (Turkey), 2013.

\bibitem{RN93}
Yiyo Kuo, Chi-Chang Wang, and Pei-Ying Chuang.
\newblock Optimizing goods assignment and the vehicle routing problem with time-dependent travel speeds.
\newblock {\em Computers \& Industrial Engineering}, 57(4):1385--1392, 2009.

\bibitem{RN91}
Chryssi Malandraki and Mark~S Daskin.
\newblock Time dependent vehicle routing problems: Formulations, properties and heuristic algorithms.
\newblock {\em Transportation science}, 26(3):185--200, 1992.

\bibitem{RN98}
Jacques Renaud, Gilbert Laporte, and Fayez~F Boctor.
\newblock A tabu search heuristic for the multi-depot vehicle routing problem.
\newblock {\em Computers \& Operations Research}, 23(3):229--235, 1996.

\bibitem{RN100}
Bin Yu, ZZ~Yang, and JX~Xie.
\newblock A parallel improved ant colony optimization for multi-depot vehicle routing problem.
\newblock {\em Journal of the Operational Research Society}, 62(1):183--188, 2011.

\bibitem{RN102}
Anand Subramanian, Puca Huachi~Vaz Penna, Eduardo Uchoa, and Luiz~Satoru Ochi.
\newblock A hybrid algorithm for the heterogeneous fleet vehicle routing problem.
\newblock {\em European Journal of Operational Research}, 221(2):285--295, 2012.

\bibitem{RN14}
Zhujun Zhang.
\newblock {\em Research on Approximation Algorithm for Heterogeneous Vehicle Routing Problem}.
\newblock Master, East China Normal University, 2014.

\bibitem{RN12}
Weizhen Rao.
\newblock {\em Research on Optimization Method for Large-Scale Dynamic Vehicle Routing Problem}.
\newblock Thesis, Dalian University of Technology, 2012.

\bibitem{RN61}
Chunshan Wang, Fan Zhang, Guifa Teng, Matthew E.Taylor, Kejian Wang, and bin Wang.
\newblock Design and implementation of smart agricultural machinery platform.
\newblock {\em Journal of Chinese Agricultural Mechanization}, 39(01):61--68, 2018.

\bibitem{RN62}
Ruyue Cao, Shichao Li, Yuhan Ji, Hongzhen Xu, Man Zhang, and Minzan Li.
\newblock Multi-machine cooperation task planning based on ant colony algorithm.
\newblock {\em Transactions of the Chinese Society for Agricultural Machinery}, 50(B07):34--39, 2019.

\bibitem{RN115}
Yiyo Kuo.
\newblock Using simulated annealing to minimize fuel consumption for the time-dependent vehicle routing problem.
\newblock {\em Computers \& Industrial Engineering}, 59(1):157--165, 2010.

\bibitem{RN16}
Meng Wang.
\newblock {\em Research on Key Technnologies on Farm Task Allocation for Multi-machine Cooperative Operation}.
\newblock Thesis, Chinese Academy of Agricultural Mechanization Sciences, 2021.

\bibitem{RN112}
Michał Okulewicz and Jacek Mańdziuk.
\newblock The impact of particular components of the pso-based algorithm solving the dynamic vehicle routing problem.
\newblock {\em Applied soft computing}, 58:586--604, 2017.

\end{thebibliography}

\end{document}